\documentclass{article}

\usepackage[dblblindworkshop, final]{neurips_2026}

\usepackage[utf8]{inputenc} 
\usepackage[T1]{fontenc}    
\usepackage{hyperref}       
\usepackage{url}            
\usepackage{booktabs}       
\usepackage{amsfonts}       
\usepackage{nicefrac}       
\usepackage{microtype}      
\usepackage{xcolor}         
\usepackage{graphicx}
\usepackage{tabularx}
\newcolumntype{Y}{>{\raggedright\arraybackslash}X}

\title{What Breaks When Facts Become Stories? The Distortions Source-Agreement Metrics Miss When Models Retell Real Events}
\workshoptitle{TAE (Trust-AI-Eval): Can We Trust AI Evaluation?}

\author{%
  Kwangsuk Park\thanks{Equal contribution.} \\
  AA Lab, Modulabs \\
  AIFFEL, Modulabs \\
  \texttt{kspark@modulabs.co.kr} \\
  \And
  Jiwoong Yang\footnotemark[1] \\
  AA Lab, Modulabs \\
  Department of Statistics, Inha University \\
  \texttt{didwldnd960923@gmail.com} \\
  \And
  Kseniia Vaniushkina\footnotemark[1] \\
  AA Lab, Modulabs \\
  AIFFEL, Modulabs \\
  \texttt{shinheepark@gmail.com} \\
}

\begin{document}

\maketitle

\begin{abstract}
Language models are now widely used to turn recorded facts into media for public consumption, yet evaluation of such output still assumes agreement with the source as its measure of success. In reconstruction, where departing from the source can itself be the goal, this criterion measures the wrong construct: applied as is, it counts intended departures as errors; withheld, it lets real distortion through. We had three model families rewrite 2,000 news articles as summaries and as stories, measured what changed on every axis of the five Ws and how, and asked of each detected distortion whether the assigned perspective warranted it, which set a third of them aside. Time and manner were not lost but replaced, while place and causation lost the expressions that carry them. Accountability distortion behaved otherwise. Barely present in summaries, it rose up to sevenfold in stories, and of those whose direction could be determined, two in three lightened the perpetrator's responsibility. Such distortions survive claim-level fact checking and leave the retelling grammatically cleaner than its source. Where our judges disagreed, three authors relabeling the same cases converged on whether a distortion had occurred but not on whether it was warranted, a boundary contested rather than merely hard to detect. To support work on detecting and reducing them, we release ReTold-2k, a reconstructable corpus with the scoring protocol and per-judge baselines kept unaggregated(https://github.com/Retold2K/retold2k).
\end{abstract}

\section{Introduction}
\label{sec:intro}
Language models are increasingly used to remake recorded facts into forms intended for public consumption. The same event may reach one reader as a short summary and another as a full story. Such reconstruction does not merely change form; it decides what to keep and what to alter. When a record is remade, what may change and what may not? The question is not confined to news. Multi-agent systems that divide such work among specialized agents face it when they rebuild a mathematics problem with its figure while keeping the original geometry intact \citep{lee2024vista}, or when they turn student code into rubric-based feedback that must keep the intent of expert graders \citep{park2025agacci}.

Whether factuality checks or overlap metrics, existing evaluation treats agreement with the source as success \citep{papineni2002bleu,lin2004rouge,pagnoni2021understanding}. For summarization, where agreement is the goal, this is a sound criterion. Where departure may be the point, difference is not failure. When generative systems turn oral-history interviews into image sequences, no agreed criterion of success exists, and scene planning conflicts with preserving the narrative in most cases \citep{park2026memory}. Applied as is, the agreement criterion counts intended departures as errors; withheld, it lets real distortion through. What this task needs is a line between departures a perspective warrants and those that misinform, the same problem that separating error from meaningful variation poses in annotation \citep{plank2022problem,sorensen2024position}, transposed to generation.

We used news as our material because the source remains verifiable and because responsibility is at stake in the events it reports. We took roughly 2,000 articles balanced across regions and topic severity and had language models rewrite each one as a summary bound to the source and as a story with the constraint relaxed. Summaries serve as the baseline condition that current evaluation has been validated on, and stories as its opposite. Following the practice of decomposing news events by the five Ws and how \citep{hamborg2019giveme5w1h}, we sorted the changes into six distortion types covering who did what to whom, why, and who said so, and for every detection we judged whether the assigned perspective warranted it. The axes left over, when, where, and how, we measured separately with rule-based indicators that involve no model judgment.

We had the models write from an assigned narrator's perspective. Changes that follow from a narrator's vantage point are what the instruction asked for, and their scope is limited to focalization, that is, who sees, what they know, and how they feel \citep{genette1980narrative}. Everything beyond that scope is what the legitimacy judgment is for, and among those changes lies a documented risk. How an event is narrated shifts where readers place responsibility \citep{entman1993framing,iyengar1991anyone,fowler1979language}. We defined twelve narrators spanning perpetrators, victims, and observers outside the event, assigned one to each article, and recorded the direction of each distortion for each narrator.

Distortion differed by axis. Time and manner were replaced by other material, and place and causation lost the expressions that carry them. Omitting substantive content and adding material absent from the source were common in both forms. What differed was accountability distortion, the kind that changes who did what to whom. It was rare in summaries and rose up to sevenfold in stories, and in two of every three cases with a determinable direction the perpetrator's responsibility was lightened. Our two model judges disagreed precisely where that responsibility was at stake. When three authors relabeled those same cases, they parted not over whether a distortion had occurred but over whether it was warranted, a sign that the boundary is contested rather than merely hard to see. Neither fact checking nor grammatical measures reach this kind of distortion, and by grammatical measures the stories read cleaner than their sources. Detecting it remains an open problem, and we release the measurement protocol and baselines as ReTold-2k to give that work a starting point.

This result is a diagnosis of evaluation method rather than a catalogue of model faults. Our contributions are as follows.

\begin{itemize}
\item We raise the problem of separating warranted departure from misinformation in reconstruction, define that boundary in judgeable form, and measure what changes across every axis of the five Ws and how.
\item We show that accountability distortion depends on the freedom the task allows, that distortions in it are lenient toward perpetrators, and that the measures in use do not detect it. Its boundary proved contested for model judges and human authors alike, and what divided them was not whether a distortion had occurred but whether it was warranted.
\item We release ReTold-2k, which bundles the judging criterion, the measurement procedure, per-judge baselines from three model families, and an evaluation set of contested cases. The contested cases carry the passages each judgment rested on and the labels of three authors.
\end{itemize}

\begin{figure}[t]
\centering
\includegraphics[width=0.85\linewidth]{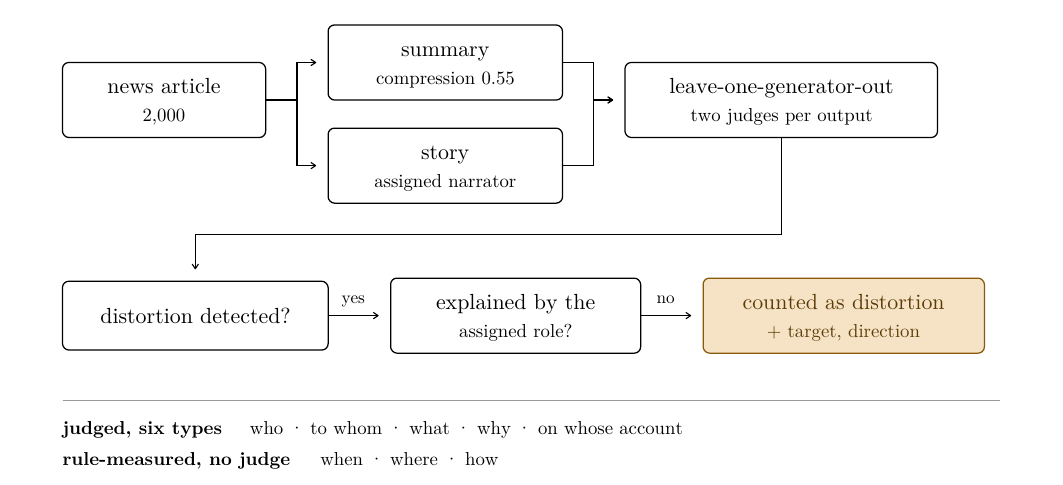}
\caption{The measurement in outline. Each article is rewritten in two forms and each output is read by the two model families that did not produce it. A change is counted only where a judge detects it and finds that the assigned role does not account for it. These are the detections we call unwarranted, and only for them are target and direction recorded. The six judged types cover five of the axes; the remaining three are measured by rule-based extractors that involve no model judgment.}
\label{fig:design}
\end{figure}

\section{Experimental Setup}
\label{sec:setup}

\paragraph{Data.}
We use news as our material because the source text remains available for comparison and because event reporting states explicitly who did what to whom. We collected 2,000 English news articles published between 2014 and 2026, stratified over seven regions and four levels of topic severity derived from event category, as summarized in Table~\ref{tab:setup} (Appendix~\ref{app:gen}). The severity strata include neutral events, in which accountability structure is sparse, so that the account that attributes distortion to the event's gravity can be tested alongside our own.

\paragraph{Two reconstruction pathways.}
Each article was rewritten in two forms that differ in how tightly they are bound to the source.

\paragraph{Summarization.}
This pathway is bound to the source and is the task on which current factuality evaluation has been validated, so it serves as the comparison condition. All three model families summarized every article at a compression ratio of 0.55; for the paired comparison we use the summary produced by the same family that generated the story. Summaries from the remaining families support per-judge baselines and robustness checks.

\paragraph{Storification.}
Each article was rewritten as a story from an assigned narrator, with model family and narrator role rotated jointly so that the twelve roles are balanced across the corpus. This yielded 1,926 outputs from 2,000 requests, the models having declined the remaining 74 (Appendix~\ref{app:decline}).

Both pathways used GPT-5.4 mini \citep{openai2026gpt54mini}, Gemini 3.1 Flash-Lite preview \citep{google2026gemini31flashlite}, and Claude Haiku 4.5 \citep{anthropic2025haiku45} uniformly, and comparisons were drawn between the two forms of the same article produced by the same model. Under this constraint 1,923 articles are paired. Assigning a single family per article reflects a design choice favoring coverage of all twelve narrator roles over cross-family comparison, and we accordingly refrain from ranking the families.

\paragraph{Judging.}
What must hold however the narrator changes is that the agent of a harmful act remains, that perpetrator and victim keep their places, that stated causes stay linked to their effects, and that contested claims remain attributed. Taking these as success conditions, we define distortion as the manner in which each is broken (Table~\ref{tab:types}). The first four conditions hold irrespective of the narrator, so inverting them isolates distortion as such; omission and addition stand on different ground, since no retelling can proceed without dropping material and supplying it. We retain the latter two because prevailing evaluation counts both as error, and we wish to see what that count distinguishes. Judging was carried out on source--summary and source--story pairs. A judge reads both texts, rules on each of the six types, and, where a distortion is detected, quotes the supporting passages from the source and from the output. Every detection is additionally judged for legitimacy, the instance-level counterpart of separating error from meaningful variation, and we record the verdict as warranted or unwarranted. A victim narrator may leave out what the victim could not have known, and that omission is warranted; deleting the agent of a harmful act is warranted by no narrator. Where accountability is implicated the target and direction are recorded as well. Because a model tends to be lenient toward its own output \citep{zheng2023judging}, we used a leave-one-generator-out protocol, in which each output was judged by the two families that did not generate it, using the same three models at temperature 0, which yielded 71,718 judgments on summaries and 23,094 on stories.

\paragraph{Rule-based indicators.}
Judging covers who, to whom, what, why, and on whose account, leaves when, where, and how untouched, and depends throughout on a model reading the text. We therefore measured every axis a second time with rule-based extractors built on spaCy \citep{honnibal2020spacy} that rely only on syntactic parsing and predefined patterns, so that no model judgment is involved. Applied alike to sources, summaries, and stories, they recover verb voice and the presence of an agent, temporal expressions, place entities, instruments of action, causal connectives, and attribution cues. We call the elements so extracted anchors, and report for each axis their density per hundred words alongside the share of source anchors that survive. What they yield is an axis map that does not depend on the judges, and cannot speak to legitimacy.

\begin{table}[t]
\centering\small
\begin{tabularx}{\linewidth}{@{}lYY@{}}
\toprule
Type & Success condition & Example of breach \\
\midrule
Vanishing Actor & A harmful act retains its agent & ``The army burned the village'' to ``the village burned'' \\
Role Reversal & Perpetrator and victim keep their places & The perpetrator is placed in the victim's position \\
Causal Fading & Stated causes stay linked to their effects & ``Died from the shelling'' to ``there was shelling, and there were deaths'' \\
Claim Laundering & Contested claims remain attributed & ``Police said'' restated as plain fact \\
Selective Omission & Who did what to whom survives & Substantive content dropped beyond what length requires \\
Invented Addition & Facts about the event come from the source & Facts or evaluations absent from the source appear \\
\bottomrule
\end{tabularx}
\caption{Six distortion types.}
\label{tab:types}
\end{table}

\section{Each axis breaks in its own way}
\label{sec:axes}
We measured the two reconstructions on every axis of the five Ws and how, once through the judges and once without them. Two axes were refilled with other material and two were thinned. The accountability axis was the exception: only there did distortion depend on the task. Vanishing Actor and Causal Fading, all but absent from summaries, rose six to sevenfold in stories.

\subsection{Time and manner are replaced, place and causation thinned}
On time and manner, density held while retention fell and novelty rose (Table~\ref{tab:rules}). These axes were not vacated but refilled. Place alone lost density with no compensating rise in novelty, so what left it was not replaced. Causation and attribution fell together, in the direction that the judges' Causal Fading and Claim Laundering also indicate. Because no model judgment enters these figures, they weigh against reading the phenomenon as an artifact of the judging procedure. Here the rule-based indicators stop. They count what has thinned; they cannot say whether the assigned perspective warrants the change. These retention and novelty figures rest on string matching and therefore miss paraphrase; they should be read alongside density.

\subsection{Freedom shifts the kind of distortion, not its amount}
\begin{figure}[t]
\centering
\includegraphics[width=0.9\linewidth]{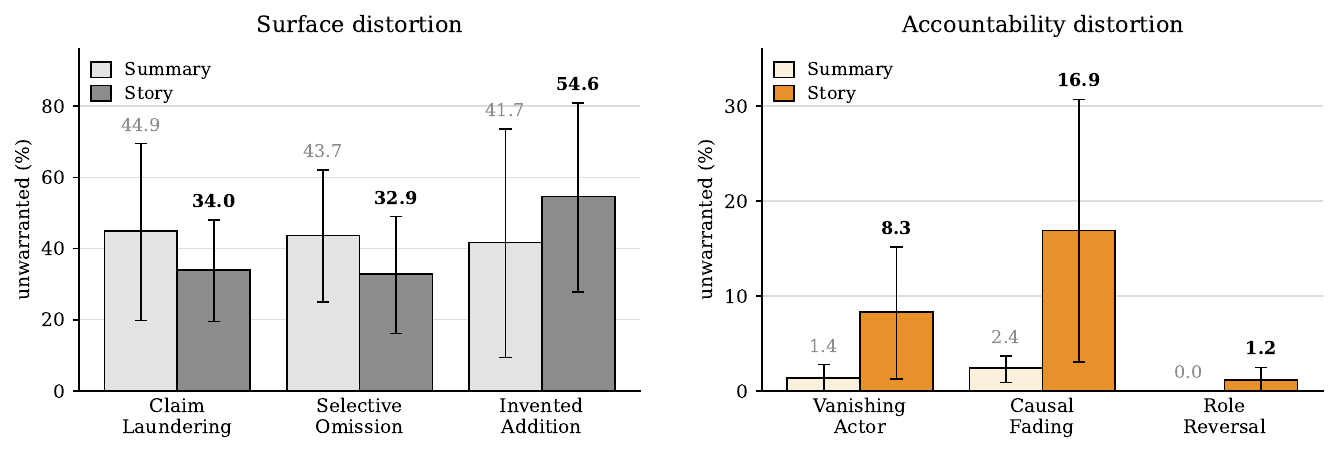}
\caption{Unwarranted detections by distortion type, summaries against stories from the same model family. Bars are means over the three judges and whiskers their standard deviation, which is wide because the judges differ in strictness. The vertical scales differ between panels, running to 88\% for surface distortion and 34\% for accountability distortion. At the article level the two tasks differ on 661 articles, 587 of them stories only (significance tests in Appendix~\ref{app:sig}). Exact figures are given in Appendix~\ref{app:num}.}
\label{fig:detect}
\end{figure}

We predicted that summarization would leave no room to alter accountability structure. Laundering attribution, dropping content, and adding material absent from the source were common in both tasks, and the first two were in fact higher in summaries (Figure~\ref{fig:detect}). The three accountability types are what separate the two, and Role Reversal did not occur in summaries at all. A third of all detections were judged warranted for the assigned narrator and are therefore excluded from the rates in Figure~\ref{fig:detect}. That excluded share varies sharply by type, exceeding half for Selective Omission and falling below four percent for Causal Fading. Widening the freedom the task allows did not raise distortion overall. It admitted a kind that had no occasion to arise in summaries. At the article level the direction was the same. Of the 661 articles on which the two tasks differed, 587 were distorted only as stories, and detections on which two judges agree stand at 6.4\% for stories against 0.4\% for summaries. Judges differ widely in strictness, yet the contrast holds within each of them.

\subsection{Task freedom, not event gravity, drives the gap}
All four strata point the same way (Table~\ref{tab:severity}). The contrast held even for neutral events, in which accountability structure is sparse. The freedom the task allows outweighs the character of the event.

\begin{table}[t]
\centering\small
\begin{tabular*}{\linewidth}{@{\extracolsep{\fill}}lccc@{}}
\toprule
Indicator & Source & Summary & Story \\
\midrule
Passive/agentless & 13.5 / 11.6 & 11.4 / 9.6 & \textbf{6.5 / 5.9} \\
Causal density & & 0.12 & \textbf{0.05} \\
Attrib. density & & 0.65 & \textbf{0.28} \\
Absolute time & & .92 / 51 / 12 & .94 / 22 / \textbf{66} \\
Instruments & & 1.13 / 15 / 63 & 1.14 / \textbf{5} / 76 \\
Place entities & & 1.89 / 56 / 15 & \textbf{1.24 / 32} / 16 \\
\bottomrule
\end{tabular*}
\caption{Rule-based indicators, applied alike to sources, summaries, and stories, with no model judgment involved. Row~1 is the share of predicates that are passive and agentless passive. The last three rows give density / retention / novelty: density is anchors per 100 words, retention the share of source anchors that survive, novelty the share of output anchors absent from the source. Read the three together: a held density with falling retention and rising novelty is replacement, whereas a falling density is loss. Where a rule-based trend runs alongside a judged type, the agreement is directional and holds at the aggregate; the two series do not necessarily flag the same items.}
\label{tab:rules}
\end{table}

\begin{table}[t]
\centering\small
\begin{tabular*}{\linewidth}{@{\extracolsep{\fill}}lcccc@{}}
\toprule
 & Neutral & Low & Medium & High \\
\midrule
Paired articles & 481 & 482 & 487 & 473 \\
Story & 23.5 & 34.6 & 37.2 & 40.0 \\
Story, both judges & 3.7 & 6.0 & 9.0 & 6.8 \\
Summary & 6.0 & 6.6 & 6.4 & 9.5 \\
\bottomrule
\end{tabular*}
\caption{Paired comparison by topic severity (\%). Article level: at least one judge detected an unwarranted accountability distortion (significance tests in Appendix~\ref{app:sig}). Neutral events are those in which accountability structure is sparse. The corresponding breakdown by region shows no difference of this kind (Appendix~\ref{app:region}).}
\label{tab:severity}
\end{table}

\section{Accountability distortion is unseen by the measures in use and tilted toward the perpetrator}
\label{sec:unseen}
The measures already in use do not see it. Grammatical indicators score the stories above their sources, and claim-level fact checking turns up nothing that contradicts the article. Yet the distortions that are detected lean one way. They lighten the perpetrator's responsibility.

\subsection{Accountability distortion leaves the grammar cleaner}
Stories carried fewer passives than the articles they retell, and fewer agentless passives (Table~\ref{tab:rules}). Rule-based indicators do not all point the same way. On causation and attribution they agreed with the judges, and on voice they run against them. By grammatical measures the story is the more transparent text, and yet unwarranted Vanishing Actor was detected six times as often as in summaries. This is the route discourse analysis has long identified, along which responsibility erodes without recourse to the passive. Table~\ref{tab:grammar} shows one instance. No clause there lacks a subject; what changes is that the subject of the assault is no longer a person. What is erased is not grammar but accountability.

\subsection{Fact checking passes accountability distortion}
Audited claim by claim against the source, 562 of the 1,445 stories in the graded strata contained nothing that contradicts it. Of those contradiction-free stories, 34.0\% carried a detected accountability distortion, and 27 of them carried one on which two judges agree. Vanishing Actor and Causal Fading operate by deletion and severance rather than addition, so they produce no sentence that conflicts with the source. Nothing in Table~\ref{tab:grammar} conflicts with its article either; only the agent is missing. The neutral stratum points the same way (Appendix~\ref{app:cr}).

\subsection{Accountability distortion tilts toward the perpetrator}
\begin{figure}[t]
\centering
\includegraphics[width=0.62\linewidth]{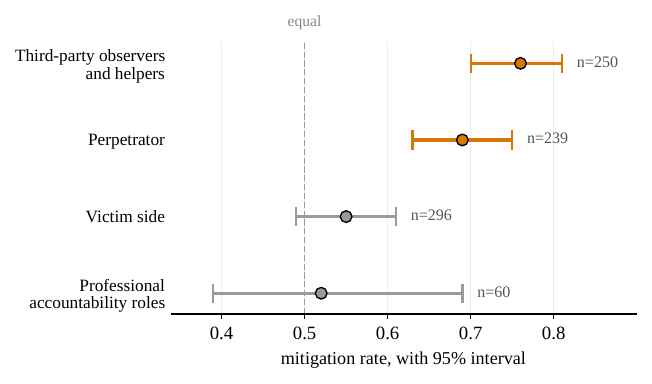}
\caption{Direction of accountability distortion toward the perpetrator, by narrator group. Points are mitigation rates with 95\% intervals, and a group counts as dominant only where the interval clears one half. \emph{n} counts detections targeting the perpetrator, including those whose direction could not be determined. Each article carries a single narrator, so narrator and article effects are not separated. The individual breakdown is given in Appendix~\ref{app:narrator}.}
\label{fig:direction}
\end{figure}

Among accountability distortions that target the perpetrator and whose direction could be determined, 67.4\% mitigated (Figure~\ref{fig:direction}). Grouping narrators by their relation to the event shows that the tilt belongs to no single voice. Observers and helpers outside the event were the most lenient, and perpetrators themselves also mitigated more often than they amplified. The narrator who erased the agent in Table~\ref{tab:grammar} was a Neutral Observer. For those close to the victim, and for the voices whose office is to assign responsibility, the interval includes one half and we withhold a verdict. The bereaved alone amplified more often than they mitigated.

\begin{table}[t]
\centering\footnotesize
\begin{tabularx}{\linewidth}{@{}p{1.8cm}Y@{}}
\toprule
\textsc{source} & Man arrested for raping ex (Cagliari, 31 Dec 2020). ``\ldots persuaded his ex-partner to come to his home \ldots and then allegedly raped her'' \\
\midrule
\textsc{story} & narrator: neutral observer, generator: GPT. ``\ldots the man invited the woman to his home under the pretext of discussing their past relationship. \ldots As the conversation progressed, it escalated into an alleged sexual assault.'' \\
\midrule
\textsc{reading} & rule-based: no passive voice, every clause carries an overt subject. judges: Vanishing Actor, detected by both, both unwarranted \\
\bottomrule
\end{tabularx}
\caption{A story that is grammatically cleaner than the article it retells. The source names an agent and an act. The story keeps active voice throughout and leaves no clause without a subject, yet the agent of the assault is displaced onto the conversation, which becomes the grammatical subject. Nothing in the passage contradicts the article; what is missing is who acted.}
\label{tab:grammar}
\end{table}

\section{The boundary divides judges and people alike}
\label{sec:boundary}
Disagreement is not evenly spread. The judges divided mainly where the perpetrator's responsibility was at stake, and three authors divided on the same cases. On whether a distortion had occurred, however, the authors converged. What divided them was whether it was warranted.

\subsection{The judges do not divide everywhere}
On accountability judgments implicating the perpetrator, pairwise disagreement reached 65.9\%; on those that did not, 2.7\% (Table~\ref{tab:disagree}). For surface distortion the perpetrator made no difference, and disagreement there exceeded a third throughout. We accordingly claim no more for the surface types than that they are common in both tasks. Where responsibility is not at stake the judges mostly agree, so this is no failure of judging competence. They divide exactly where the judgment is hard, at the point where one must ask whether a change is the narrator's prerogative or misinformation.

\begin{table}[t]
\centering\small
\begin{tabular*}{0.8\linewidth}{@{\extracolsep{\fill}}lcc@{}}
\toprule
 & Perp. implicated & Other \\
\midrule
Accountability distortion & \textbf{65.9 $\pm$ 18.9} & 2.7 $\pm$ 1.4 \\
Surface distortion & 36.1 $\pm$ 17.3 & 35.3 $\pm$ 11.8 \\
\bottomrule
\end{tabular*}
\caption{Judge disagreement by whether the perpetrator is implicated (\%, mean $\pm$ s.d. over three judge pairs). Pooled figures and the significance test are given in Appendix~\ref{app:sig}.}
\label{tab:disagree}
\end{table}

\subsection{People divide in the same place}
Three authors independently relabeled the 50 cases on which the judges had divided. Where the judges had divided over whether a distortion occurred, four cases in ten drew unanimous labels; where they had divided over legitimacy, one in ten did. Fleiss' $\kappa$ \citep{fleiss1971measuring}, which measures inter-rater agreement corrected for chance (1 is perfect agreement, 0 is chance level), was $-0.07$ on the legitimacy split, below chance (Table~\ref{tab:human}a). The individual labels show why (Table~\ref{tab:human}b). One author read almost all eighteen as the narrator's prerogative, while another read most of them as accountability distortion. The spread in strictness among the model judges recurs among people. The most permissive author and the strictest share a first language, and the author who worked in a different one falls between them, so neither language nor translation accounts for the spread.

\subsection{We do not aggregate}
In eight of the fifty cases the three answers admit no majority. Collapsing them to a single label would discard that fact. Treating annotator disagreement as signal about the object rather than as noise is established in work on labeling \citep{plank2022problem,cabitza2023toward}, and the same structure appears in judgments of generated text. In the evaluation of oral-history visualization, the conflict between scene planning and narrative preservation was invisible under aggregated scoring and visible only when values were scored separately \citep{park2026memory}. We therefore release the judgments unaggregated, judge by judge (Section~\ref{sec:retold}).

\begin{table}[t]
\centering\small
\begin{minipage}[t]{0.58\linewidth}
\centering
\emph{(a) Author agreement, by the layer at which the judges divided}\\[3pt]
\begin{tabular}{@{}p{3.4cm}rrrr@{}}
\toprule
Judges divided over & n & Unan. & Tied & $\kappa$ \\
\midrule
whether a distortion occurred & 32 & 13 & 2 & 0.38 \\
whether it was warranted & 18 & 2 & 6 & $\mathbf{-0.07}$ \\
all cases & 50 & 15 & 8 & 0.23 \\
\bottomrule
\end{tabular}
\end{minipage}\hfill
\begin{minipage}[t]{0.38\linewidth}
\centering
\emph{(b) Individual labels on the 18 legitimacy-split cases}\\[3pt]
\begin{tabular}{@{}lrrr@{}}
\toprule
Author & Warr. & Not warr. & No dist. \\
\midrule
A1 & 15 & 0 & 3 \\
A2 & 2 & 12 & 4 \\
A3 & 5 & 5 & 8 \\
\bottomrule
\end{tabular}
\end{minipage}
\caption{Three authors independently relabeled 50 cases on which the model judges had divided, each directed to the passages a judge had flagged without seeing the judges' verdicts or one another's labels, choosing among warranted, not warranted, and no distortion present. Unanimous and Tied are counts out of n. All 18 cases of legitimacy disagreement were taken, and the 32 detection-disagreement cases were sampled stratified by topic severity and distortion type. The 50 were drawn solely from judgments on which the judges divided; they are not a random sample, and $\kappa$ here measures how contested the boundary is, not how reliable the judgments are overall.}
\label{tab:human}
\end{table}

\section{ReTold-2k}
\label{sec:retold}
Aggregation removes the one trace this distortion reliably leaves: that judges divide over it. ReTold-2k keeps that trace. It comprises the criterion that settles which departures an assigned narrator warrants, the procedure that applies that criterion to any output, the per-judge baselines left by three model families, and an evaluation set assembled from the cases on which the judges divided (Table~\ref{tab:release}).

Records are kept judge by judge. Each carries the detection and the legitimacy verdict and, where accountability is implicated, the target, the direction, and the passages from source and output on which the judgment rested, quoted verbatim. The two records for an item are left side by side to the end; Appendix~\ref{app:record} shows a worked example (Table~\ref{tab:record}). Our own outputs serve as points of comparison rather than as test items for another model, so the article list and the narrator assignment rule accompany them, with the prompts reproduced in full in Appendices~\ref{app:gen} and~\ref{app:judge}.

What functions as a benchmark is the set of 911 contested judgments, to which the individual labels of the three authors who relabeled fifty of them are attached. Since having no settled answer is what defines this subset, asking how closely a system matches a majority-vote gold loses its meaning here. We propose instead that agreement be reported against each judge taken separately, together with the spread across them.

Rerunning the procedure will not, however, yield figures directly comparable to ours, since absolute levels shift with the judge models employed. That is why the per-judge labels and the prompts are released in full, and why the rule-based indicators, which pass through no judge, remain comparable over time.

\begin{table}[t]
\centering\footnotesize
\begin{tabularx}{\linewidth}{@{}p{0.42\linewidth}Y@{}}
\toprule
Released & Why it is needed \\
\midrule
Article index: URL, date, region, severity, title (2,000) & Rebuilds the same corpus. Copyright bars the article text itself \\
\addlinespace
Collection script, prompts, narrator definitions and assignment rule, extractors, aggregation script & A changed prompt is a changed task. Without these the numbers are not comparable \\
\addlinespace
Summaries at compression 0.55 (5,985) and stories (1,926) & Fixed inputs for anyone measuring a detector rather than a generator \\
\addlinespace
Declined requests with the model's reply (74) & What the task fails to elicit belongs to the measurement \\
\addlinespace
Judgments, per judge, unaggregated (94,812) & Our baseline, and the labels against which a detector is scored \\
\addlinespace
Contested judgments with source and output passages quoted verbatim (911) & The hard cases must stay usable after links rot \\
\addlinespace
Author labels, kept individual (50 $\times$ 3) & No aggregated gold exists for this subset \\
\bottomrule
\end{tabularx}
\caption{What ReTold-2k releases, and why each part is needed. Article text, the summaries at the other two compression ratios, and the judgments on them are not included. The paper's claims do not rest on them, and shipping unused runs only invites confusion with the reported figures.}
\label{tab:release}
\end{table}

\section{Conclusion}
\label{sec:disc}
Evaluation of reconstruction has to measure not agreement with the source but whether each departure is warranted. We defined that criterion in judgeable form and applied it to every detection, and what it revealed is a distortion that depends on the task. Accountability distortion, the kind that changes who did what to whom, was all but absent from summaries and rose six to sevenfold in stories, and of the cases whose direction could be determined, two in three lightened the perpetrator's responsibility. The measures in use do not see it. One in three stories that passed claim-level fact checking carried an accountability distortion, and over the full set the grammatical indicators scored the stories above their sources. What separated the tasks was not the gravity of the event but the latitude the task allowed. Verification therefore has to track how loose the task is.

Whether a departure occurred and whether it was warranted are different questions. Model judges and human authors alike converged on the first and parted on the second, and the parting concentrated on cases where the perpetrator's responsibility was at stake. As long as the boundary between warranted and unwarranted remains contested, aggregating the judgments into a single figure erases the very thing in dispute. That is why ReTold-2k keeps the judgments unaggregated, judge by judge. What counts as an error depends on who is judging. A measure that returns one number has already decided that question for us.

\bibliographystyle{plainnat}
{\small
\bibliography{references}

@inproceedings{hamborg2019giveme5w1h,
  title     = {Giveme5W1H: A Universal System for Extracting Main Events from News Articles},
  author    = {Hamborg, Felix and Breitinger, Corinna and Gipp, Bela},
  booktitle = {Proceedings of the 13th ACM Conference on Recommender Systems,
               7th International Workshop on News Recommendation and Analytics (INRA)},
  year      = {2019},
  address   = {Copenhagen, Denmark}
}

@article{entman1993framing,
  title   = {Framing: Toward Clarification of a Fractured Paradigm},
  author  = {Entman, Robert M.},
  journal = {Journal of Communication},
  volume  = {43}, number = {4}, pages = {51--58}, year = {1993}, month = dec,
  publisher = {Oxford University Press},
  doi     = {10.1111/j.1460-2466.1993.tb01304.x}
}

@book{iyengar1991anyone,
  title     = {Is Anyone Responsible? How Television Frames Political Issues},
  author    = {Iyengar, Shanto},
  publisher = {University of Chicago Press}, address = {Chicago},
  year      = {1991}
}

@book{fowler1979language,
  title     = {Language and Control},
  author    = {Fowler, Roger and Hodge, Bob and Kress, Gunther and Trew, Tony},
  publisher = {Routledge and Kegan Paul}, address = {London},
  year      = {1979}
}

@book{genette1980narrative,
  title       = {Narrative Discourse: An Essay in Method},
  author      = {Genette, G{\'e}rard},
  translator  = {Lewin, Jane E.},
  publisher   = {Cornell University Press}, address = {Ithaca, NY},
  year        = {1980},
  isbn        = {978-0-8014-9259-4}
}

@inproceedings{plank2022problem,
  title     = {The ``Problem'' of Human Label Variation: On Ground Truth in Data, Modeling and Evaluation},
  author    = {Plank, Barbara},
  booktitle = {Proceedings of the 2022 Conference on Empirical Methods in Natural Language Processing},
  month = dec, year = {2022}, address = {Abu Dhabi, United Arab Emirates},
  publisher = {Association for Computational Linguistics},
  pages     = {10671--10682}, doi = {10.18653/v1/2022.emnlp-main.731}
}

@inproceedings{cabitza2023toward,
  title     = {Toward a Perspectivist Turn in Ground Truthing for Predictive Computing},
  author    = {Cabitza, Federico and Campagner, Andrea and Basile, Valerio},
  booktitle = {Proceedings of the 37th AAAI Conference on Artificial Intelligence},
  year      = {2023}
}

@inproceedings{sorensen2024position,
  title     = {Position: A Roadmap to Pluralistic Alignment},
  author    = {Sorensen, Taylor and Moore, Jared and Fisher, Jillian and Gordon, Mitchell
               and Mireshghallah, Niloofar and Rytting, Christopher Michael and Ye, Andre
               and Jiang, Liwei and Lu, Ximing and Dziri, Nouha and Althoff, Tim and Choi, Yejin},
  booktitle = {Proceedings of the 41st International Conference on Machine Learning (ICML)},
  year      = {2024}
}

@inproceedings{pagnoni2021understanding,
  title     = {Understanding Factuality in Abstractive Summarization with FRANK: A Benchmark for Factuality Metrics},
  author    = {Pagnoni, Artidoro and Balachandran, Vidhisha and Tsvetkov, Yulia},
  booktitle = {Proceedings of the 2021 Conference of the North American Chapter of the
               Association for Computational Linguistics: Human Language Technologies},
  month = jun, year = {2021}, address = {Online},
  publisher = {Association for Computational Linguistics},
  pages     = {4812--4829}, doi = {10.18653/v1/2021.naacl-main.383}
}

@inproceedings{zheng2023judging,
  title     = {Judging LLM-as-a-Judge with MT-Bench and Chatbot Arena},
  author    = {Zheng, Lianmin and Chiang, Wei-Lin and Sheng, Ying and Zhuang, Siyuan
               and Wu, Zhanghao and Zhuang, Yonghao and Lin, Zi and Li, Zhuohan
               and Li, Dacheng and Xing, Eric P. and Zhang, Hao and Gonzalez, Joseph E. and Stoica, Ion},
  booktitle = {Advances in Neural Information Processing Systems 36,
               Datasets and Benchmarks Track},
  year      = {2023}
}

@article{fleiss1971measuring,
  title   = {Measuring Nominal Scale Agreement Among Many Raters},
  author  = {Fleiss, Joseph L.},
  journal = {Psychological Bulletin}, volume = {76}, number = {5},
  pages   = {378--382}, year = {1971}, month = nov,
  doi     = {10.1037/h0031619}
}

@article{mcnemar1947note,
  title   = {Note on the Sampling Error of the Difference Between Correlated Proportions or Percentages},
  author  = {McNemar, Quinn},
  journal = {Psychometrika}, volume = {12}, number = {2},
  pages   = {153--157}, year = {1947}, month = jun,
  doi     = {10.1007/BF02295996}
}

@inproceedings{lin2004rouge,
  title     = {ROUGE: A Package for Automatic Evaluation of Summaries},
  author    = {Lin, Chin-Yew},
  booktitle = {Text Summarization Branches Out},
  month = jul, year = {2004}, address = {Barcelona, Spain},
  publisher = {Association for Computational Linguistics},
  pages     = {74--81}
}

@inproceedings{papineni2002bleu,
  title     = {BLEU: A Method for Automatic Evaluation of Machine Translation},
  author    = {Papineni, Kishore and Roukos, Salim and Ward, Todd and Zhu, Wei-Jing},
  booktitle = {Proceedings of the 40th Annual Meeting of the Association for Computational Linguistics},
  month = jul, year = {2002}, address = {Philadelphia, Pennsylvania, USA},
  publisher = {Association for Computational Linguistics},
  pages     = {311--318}, doi = {10.3115/1073083.1073135}
}

@misc{honnibal2020spacy,
  title  = {spaCy: Industrial-strength Natural Language Processing in Python},
  author = {Honnibal, Matthew and Montani, Ines and Van Landeghem, Sofie and Boyd, Adriane},
  publisher = {Zenodo},
  year   = {2020}, doi = {10.5281/zenodo.1212303}
}

@misc{openai2026gpt54mini,
  title  = {Introducing GPT-5.4 mini and nano},
  author = {{OpenAI}}, year = {2026}, month = mar,
  url    = {https://openai.com/index/introducing-gpt-5-4-mini-and-nano/}
}

@misc{google2026gemini31flashlite,
  title  = {Gemini 3.1 Flash-Lite: Model Card},
  author = {{Google DeepMind}}, year = {2026}, month = mar,
  url    = {https://deepmind.google/models/model-cards/gemini-3-1-flash-lite/}
}

@misc{anthropic2025haiku45,
  title  = {Introducing Claude Haiku 4.5},
  author = {{Anthropic}}, year = {2025}, month = oct,
  url    = {https://www.anthropic.com/news/claude-haiku-4-5}
}

@article{lee2024vista,
  title   = {{VISTA}: Visual Integrated System for Tailored Automation in Math Problem Generation Using {LLM}},
  author  = {Lee, Jeongwoo and Park, Kwangsuk and Park, Jihyeon},
  journal = {arXiv preprint arXiv:2411.05423},
  year    = {2024}
}

@article{park2025agacci,
  title   = {{AGACCI}: Affiliated Grading Agents for Criteria-Centric Interface in Educational Coding Contexts},
  author  = {Park, Kwangsuk and Yang, Jiwoong},
  journal = {arXiv preprint arXiv:2507.05321},
  year    = {2025}
}

@article{park2026memory,
  title   = {What Gets Lost When Memory Becomes Media? {E}valuating {AI}-Generated Oral History Visualization},
  author  = {Park, Kwangsuk and Koo, Jaehyun and Lee, Jiyeon and Tan, Anjung and Park, Hyoungchul},
  journal = {arXiv preprint arXiv:2607.24756},
  year    = {2026}
}
}


\appendix

\section{Data and setup}
\label{app:gen}
Table~\ref{tab:setup} lists the strata used to build the corpus: seven regions, four levels of topic severity, and the twelve narrator roles grouped by their relation to the event. Region and severity are assigned per article; one narrator is assigned to each article as described below.

\begin{table}[htbp]
\centering\small
\begin{tabularx}{\linewidth}{@{}llY@{}}
\toprule
Dimension & Subgroup & Examples \\
\midrule
Region & North America & US, Canada \\
 & South America & Brazil, Venezuela \\
 & Europe & UK, France, Ukraine \\
 & Russia \& CIS & Russia, Kazakhstan \\
 & Middle East & Iraq, Iran, Israel \\
 & Africa & South Africa, Nigeria \\
 & Asia \& Oceania & China, Japan, Australia \\
\midrule
Severity & High & terrorist attack, massacre \\
 & Medium & homicide, airstrike \\
 & Low & disaster, financial fraud \\
 & Neutral & diplomacy, trade, sports \\
\midrule
Narrator & Key personnel & victim, survivor, perpetrator, rescuer \\
 & Relations & witness, bereaved family, neighbor, descendant \\
 & Observers & journalist, historian, investigator, neutral observer \\
\bottomrule
\end{tabularx}
\caption{Experimental setup: regions, topic severity, and narrator classes.}
\label{tab:setup}
\end{table}

\section{Generation}
\label{app:generation}

\paragraph{Storification prompt.}
Figure~\ref{fig:prompt-story} reproduces the prompt as used.

\paragraph{Summarization.}
The same clients produced summaries at a compression ratio of 0.55. No narrator is assigned to a summary.

\paragraph{Narrators.}
Twelve roles, grouped by their relation to the event. Third-party observers and helpers: Witness, Neutral Observer, Rescuer, Neighbor, Historian. Perpetrator: Perpetrator. Victim side: Victim, Survivor, Descendant, Bereaved family. Professional accountability roles: Journalist, Investigator. Articles were shuffled with seed 42, then model family and narrator were rotated jointly so that both are balanced across the corpus.

\paragraph{Two notes on the prompt.}
The phrases \emph{historical story} and \emph{illustrated in 4 storyboard images} are inherited from the earlier project for which this corpus was first built. No images are generated here, and these phrases act only as a demand for concrete detail. We left the prompt unchanged so as not to alter the generation condition. The prompt also asks for roughly 200 words, which is why the stories average 194 words, shorter than the summaries they are paired with at 304.

\begin{figure}[htbp]
\begin{small}
\begin{verbatim}
You are a "Perspective Narrative Agent".

GOAL:
Rewrite the following historical story from the specific
perspective of a "{perspective}".

ORIGINAL STORY:
<<<
{original_story}
>>>

CONDITION:
1. POV: Use the voice and emotional tone of a "{perspective}".
2. Timeline: Start the story with the "YYYY-MM" format based on
   the source's date.
3. Length: Approximately 200 words.
4. Prefix:
   - Start with "Y" if you can successfully rewrite it into a story.
   - Start with "N" if the source content is insufficient to create
     a 200-word story.

CONSTRAINTS:
- If "Y": Output format must be: Y \n [YYYY-MM], [Rewritten Story]
- If "N": Output format must be: N \n [Reason]
- Maintain chronological order and do not invent facts.
- The style should be vivid enough to be illustrated in 4
  storyboard images.

Output:
\end{verbatim}
\end{small}
\caption{The storification prompt as used, reproduced verbatim. The variables are \texttt{\{perspective\}} and \texttt{\{original\_story\}}.}
\label{fig:prompt-story}
\end{figure}

\section{Judging}
\label{app:judge}

\paragraph{Judging prompt.}
Figures~\ref{fig:prompt-judge1}, \ref{fig:prompt-judge2} and~\ref{fig:prompt-judge3} reproduce the prompt as used. The source, the output, and the assigned \texttt{PERSPECTIVE} are supplied with it.

\paragraph{How summaries are treated.}
As the prompt states, when \texttt{PERSPECTIVE} is \texttt{None (faithful summary)} no departure is licensed, so every detection in a summary is recorded as unwarranted. A summary has no narrator and therefore no perspective by which a departure could be warranted. This asymmetry makes our comparison conservative. In Figure~\ref{fig:detect} the summary columns carry every detection, while the story columns carry only what remains after a third is set aside as warranted. Accountability distortion is six to sevenfold higher in stories even so.

\paragraph{Contradiction-rate audit.}
The NLI-style audit prompt was carried over unchanged from the earlier project; we verified that prompt and model are character-identical to the original. All generation and judging ran through the providers' commercial APIs; no models were trained, and the rule-based extraction runs on a single CPU workstation.

\begin{figure}[htbp]
\begin{small}
\begin{verbatim}
Compare a news article (source) with a RETELLING of the same event,
written in the voice of an assigned narrator perspective (given as
PERSPECTIVE). Unlike a summary, a retelling is allowed to change
style, voice, order, emotional coloring, and to omit or compress
details the narrator would plausibly not know or dwell on. Apply
each definition below LITERALLY to decide detected(0/1); then,
separately, judge licensed(0/1): whether that deviation is a natural
consequence of narrating from PERSPECTIVE (licensed=1) or would
misinform any reader about what actually happened and who is
responsible, regardless of who is telling it (licensed=0).

Hard rule: changes to the event's responsibility structure - who did
what to whom, and what caused what - are NEVER licensed by any
perspective.

If PERSPECTIVE is 'None (faithful summary)', treat the text as a
summary: no deviation is licensed by perspective (licensed=0
whenever detected=1).
\end{verbatim}
\end{small}
\caption{The judging prompt, part one: the comparison the judge is asked to make, the separation of detection from legitimacy, and the two rules that bound it.}
\label{fig:prompt-judge1}
\end{figure}

\begin{figure}[htbp]
\begin{small}
\begin{verbatim}
Definitions:
- AR agency removal: a harmful action whose doer is explicit in the
  source becomes agentless or passive in the retelling (the doer
  disappears from that action).
- RI relation inversion: perpetrator and victim roles are swapped or
  confused relative to the source.
- AS attribution stripping: a claim the source attributes to a
  speaker or source (e.g., 'police said', 'according to X') is
  stated as plain fact in the retelling - claim retained,
  attribution gone. (First-person absorption of attribution into
  lived experience may be licensed UNLESS it converts a contested
  claim into fact or changes who is claimed responsible.)
- CD causal dilution: an explicit cause-effect link in the source is
  weakened to mere co-occurrence or sequence in the retelling while
  both events are still mentioned.
- SO selective omission: substantive content units of the source
  (especially who-did-what-to-whom) are absent from the retelling
  beyond what its length requires. (Omission of details the narrator
  could not know may be licensed UNLESS it removes the perpetrator's
  role or the victim's harm.)
- IJ information injection: NEW facts, events, attributions, or
  evaluative framings present in the retelling but absent from the
  source, including demographic attributes of persons. (The
  narrator's own inner states, feelings, and generic scene color may
  be licensed; new factual claims about the event, participants,
  causes, or blame are not.)

Do NOT flag mere paraphrase or lexical generalization.
\end{verbatim}
\end{small}
\caption{The judging prompt, part two: the definition of each of the six distortion types.}
\label{fig:prompt-judge2}
\end{figure}

\begin{figure}[htbp]
\begin{small}
\begin{verbatim}
Judge each code independently; detecting one does not preclude
others.

For each distortion, output one pipe-delimited line:
CODE | detected(0/1) | licensed(0/1) | target_role | position |
valence | evidence
- licensed: 1 only if detected=1 AND the deviation is licensed by
  PERSPECTIVE as defined above; 0 if detected=1 and not licensed;
  na if detected=0
- target_role: perpetrator / victim / neutral / na
- position: lead / body / na
- valence: mitigate / amplify / neutral
- evidence: brief source->retelling, empty if not detected

Output exactly 6 lines (AR,RI,AS,CD,SO,IJ), no header.
\end{verbatim}
\end{small}
\caption{The judging prompt, part three: how each judgment is to be reported.}
\label{fig:prompt-judge3}
\end{figure}

\section{Record format}
\label{app:record}
Each item in ReTold-2k is stored as two records, one per judging family, and the two are kept side by side rather than resolved into a single verdict. Table~\ref{tab:record} shows one such item: both judges detect the same distortion type and call it unwarranted, yet assign different targets and directions, which is exactly the disagreement the resource is meant to preserve. Evidence fields are quoted verbatim from the judges' output.

\begin{table}[htbp]
\centering\footnotesize
\begin{tabularx}{\linewidth}{@{}p{2.2cm}Y@{}}
\toprule
\multicolumn{2}{@{}l}{item \texttt{NN\_gemini\_0447\_Victim}, type Causal Fading}\\
\multicolumn{2}{@{}l}{story, narrator: victim, generator: Gemini}\\
\midrule
\multicolumn{2}{@{}l}{\textsc{judge = claude}}\\
detected & true \\
warranted & false \\
target / dir. & perpetrator / amplify \\
evidence & ``Source: explicit causal chain, Retadup malware to cryptocurrency creation and ransomware. Retelling: presents ransomware demand as primary harm without establishing that Retadup was the vector\ldots'' \\
\midrule
\multicolumn{2}{@{}l}{\textsc{judge = gpt}}\\
detected & true \\
warranted & false \\
target / dir. & neutral / neutral \\
evidence & ``source: police removed malware after counterattack; retelling: `They even managed to remove the virus!' with no causal chain to the police action'' \\
\bottomrule
\end{tabularx}
\caption{One record of ReTold-2k, drawn from the contested subset. Each output is judged by the two families that did not generate it, and the two records are stored side by side rather than resolved. Both judges detect Causal Fading and call it unwarranted, yet they assign different targets and directions. Evidence fields are reproduced verbatim.}
\label{tab:record}
\end{table}

\section{Contradiction-rate audit, by stratum}
\label{app:cr}
We audited the 481 neutral-stratum stories by the same procedure. 209 of them (43.5\%) contain nothing that contradicts the source article, and 24.9\% of those contradiction-free stories carry a detected accountability distortion, 3 of them on which two judges agree. This runs in the same direction as the graded strata reported in Section~\ref{sec:unseen}, where 562 of 1,445 stories were contradiction-free and 34.0\% of those carried a detected distortion. Absolute levels are lower for neutral events, in which accountability structure is sparse, but the blind spot does not disappear.

\section{Regional breakdown}
\label{app:region}
The corpus is stratified over seven regions: Africa 251, Asia and Oceania 306, Europe 296, Middle East 268, North America 374, Russia and CIS 246, South America 259. Rule-based indicators broken down by region are released with the resource. Differences between regions in retention and novelty are small relative to differences between severity strata and show no consistent direction. On that basis the paper claims only that region does not account for the contrast.

\section{Declined requests}
\label{app:decline}
The models declined 74 of the 2,000 storification requests. The classification below matches surface cues in the refusal text and has not been checked by the authors: policy grounds 42, no event to narrate 23, insufficient information 6, no rule matched 3. Refusals are concentrated in one family. Claude accounts for 72, GPT for 2, and Gemini for none. Because each article is assigned to a single family, refusals therefore fall on particular articles rather than spreading evenly. By stratum, 57 come from the graded set and 17 from the neutral set. A declined request leaves no output to judge and is excluded from the detection rates. We did not judge whether declining was warranted; the same criterion could be asked of it.

\section{Direction by individual narrator}
\label{app:narrator}
Seven narrators mitigate more often than they amplify, four are undetermined, and one amplifies. The most lenient narrator is the disinterested Neutral Observer; the only one to amplify is the bereaved family member.

\begin{table}[htbp]
\centering\small
\begin{tabular*}{0.85\linewidth}{@{\extracolsep{\fill}}lrrrrc@{}}
\toprule
Narrator & n & Mit. & Amp. & Rate & 95\% CI \\
\midrule
Neutral Observer & 42 & 35 & 2 & .83 & .72--.95 \\
Rescuer & 82 & 67 & 14 & .82 & .73--.90 \\
Survivor & 44 & 33 & 10 & .75 & .62--.88 \\
Descendant & 49 & 35 & 12 & .71 & .59--.84 \\
Historian & 48 & 34 & 8 & .71 & .58--.84 \\
Neighbor & 43 & 30 & 12 & .70 & .56--.83 \\
Perpetrator & 239 & 165 & 72 & .69 & .63--.75 \\
\midrule
Witness & 35 & 23 & 8 & .66 & .50--.81 \\
Journalist & 37 & 21 & 14 & .57 & .41--.73 \\
Victim & 124 & 65 & 58 & .52 & .44--.61 \\
Investigator & 23 & 10 & 11 & .43 & .23--.64 \\
\midrule
Bereaved family & 79 & 30 & 44 & .38 & .27--.49 \\
\bottomrule
\end{tabular*}
\caption{Direction by individual narrator. n counts detections that target the perpetrator and were not warranted by the assigned perspective; the rate is mitigating cases over n; intervals are normal-approximation 95\%. The upper block mitigates, the middle is undetermined, and the lower amplifies.}
\label{tab:narrator}
\end{table}

\section{Significance tests}
\label{app:sig}
We tested whether the paired task contrasts in the body could have arisen by chance. At the article level we paired, for each article, whether an unwarranted accountability distortion was detected in the summary and in the story, and applied McNemar's exact two-sided test \citep{mcnemar1947note}. Over all paired articles the difference is significant at $p<10^{-99}$ (Figure~\ref{fig:detect}). Broken down by topic severity, the test gives $p<10^{-15}$ for neutral events, $10^{-29}$ for low, $10^{-33}$ for medium, and $10^{-27}$ for high severity (Table~\ref{tab:severity}). Separately, disagreement among the judges depends on whether the perpetrator is implicated: pooling all judgments, the rate is 58.8\% when the perpetrator is implicated against 3.1\% when not, a difference significant under a $\chi^2$ test at $p<10^{-200}$ (Table~\ref{tab:disagree}).

\section{Exact figures for Figure~\ref{fig:detect}}
\label{app:num}

\begin{table}[htbp]
\centering\small
\begin{tabular*}{0.75\linewidth}{@{\extracolsep{\fill}}lcc@{}}
\toprule
Type & Summary & Story \\
\midrule
Claim Laundering & 44.9 $\pm$ 24.5 & 34.0 $\pm$ 13.8 \\
Selective Omission & 43.7 $\pm$ 18.1 & 32.9 $\pm$ 16.1 \\
Invented Addition & 41.7 $\pm$ 31.7 & 54.6 $\pm$ 26.2 \\
Vanishing Actor & 1.4 $\pm$ 1.3 & 8.3 $\pm$ 6.8 \\
Causal Fading & 2.4 $\pm$ 1.3 & 16.9 $\pm$ 13.6 \\
Role Reversal & 0.0 & 1.2 $\pm$ 1.3 \\
\bottomrule
\end{tabular*}
\caption{Unwarranted detection rates by task (\%, mean $\pm$ s.d. over three judges). Summaries are those produced by the same family that generated the story, at compression 0.55. The $\pm$ spans wide differences in judge strictness rather than run-to-run variation; comparing each judge against that judge's own summary baseline yields a ratio of at least 3.4 in all nine cells.}
\label{tab:exact}
\end{table}


\end{document}